\documentclass[11pt]{article}
\usepackage{amsmath,amsthm,amsfonts,amssymb}
\usepackage[margin=1in]{geometry}
\usepackage{fancyhdr}
\usepackage{xcolor}
\usepackage{tcolorbox}
\usepackage{graphicx}
\usepackage{subcaption}
\usepackage{tcolorbox}
\usepackage{enumerate}
\usepackage{tikz}
\usepackage{hyperref}
\usepackage{multicol}
\usepackage{enumitem}
\usepackage{makecell}
\usepackage{tabularx}

\title{\vspace{-2.0cm}\textbf{LLM-ARC: Enhancing LLMs with an Automated Reasoning Critic}}

\author{
Aditya Kalyanpur\footnote{Corresponding Author: adityak@ec.ai}, Kailash Karthik Saravanakumar, Victor Barres,\\
Jennifer Chu-Carroll, David Melville, David Ferrucci\\
\\
Elemental Cognition Inc.\\
}

\begin{document}

\date{} 

\maketitle

\begin{abstract}
We introduce LLM-ARC, a neuro-symbolic framework designed to enhance the logical reasoning capabilities of Large Language Models (LLMs), by combining them with an Automated Reasoning Critic (ARC). LLM-ARC employs an Actor-Critic method where the LLM Actor generates declarative logic programs along with tests for semantic correctness, while the Automated Reasoning Critic evaluates the code, runs the tests and provides feedback on test failures for iterative refinement. Implemented using Answer Set Programming (ASP), LLM-ARC achieves a new state-of-the-art accuracy of 88.32\% on the FOLIO benchmark which tests complex logical reasoning capabilities. Our experiments demonstrate significant improvements over LLM-only baselines, highlighting the importance of logic test generation and iterative self-refinement. We achieve our best result using a fully automated self-supervised training loop where the Actor is trained on end-to-end dialog traces with Critic feedback. We discuss potential enhancements and provide a detailed error analysis, showcasing the robustness and efficacy of LLM-ARC for complex natural language reasoning tasks.
\end{abstract}

\section{Introduction}

\begin{multicols}{2}
Given their impressive language understanding capability, Large Language Models (LLMs) are being used to develop a wide variety of Natural Language applications. For certain classes of applications that require a high degree of accuracy and reliability (e.g., enterprise applications in the medical, legal or finance domain), LLMs are often combined with external tools and solvers in a hybrid architecture \cite{gur2024realworld, schick2023toolformer, liang2023taskmatrixai}. We believe this is the right approach, especially to tackle problems where precise logical reasoning, planning or constraint optimization is required, as LLMs are known to struggle for this class of problems \cite{zheng2024natural, valmeekam2023planning, kambhampati2024llms, chucarroll2024llms}. 

In this work, we focus on logical reasoning problems expressed in natural language, for which there has been a growing interest in developing neuro-symbolic architectures \cite{linc, logiclm}. These architectures combine the power of LLMs for generating (declarative) code and filling in missing background (commonsense) knowledge, with the accuracy of automated Symbolic Reasoning systems to do precise logical reasoning. 

This design addresses the limitations of either technology when used independently: the LLMs' inability to do accurate and consistent reasoning based on the underlying domain logic, and the Symbolic Reasoner's inability to work with unstructured data, and explicitly encode common-sense knowledge to get the desired inferences. The former issue in symbolic systems is the well-known ``knowledge acquisition" problem, while the latter issue typically leads to their brittleness.

Building on our previous neuro-symbolic work \cite{kalyanpur2021braid, chucarroll2024llms}, we develop a new framework based on the \emph{Actor-Critic} \cite{actor-critic} model, where the \textbf{Actor} generates declarative code, crucially with tests to verify the semantic correctness of the code (i.e. the logic program correctly captures the modeler's intent), and the \textbf{Critic} runs the code and tests, and gives feedback with detailed explanations to the Actor if the code does not compile or some tests fail. When this happens, the Actor re-generates the code/tests based on the feedback, and the process is repeated with the Critic evaluating the results until all tests pass, or we reach a max-iteration limit. 
 
\begin{figure*}[htb]
    \centering
    \includegraphics[width=\textwidth]{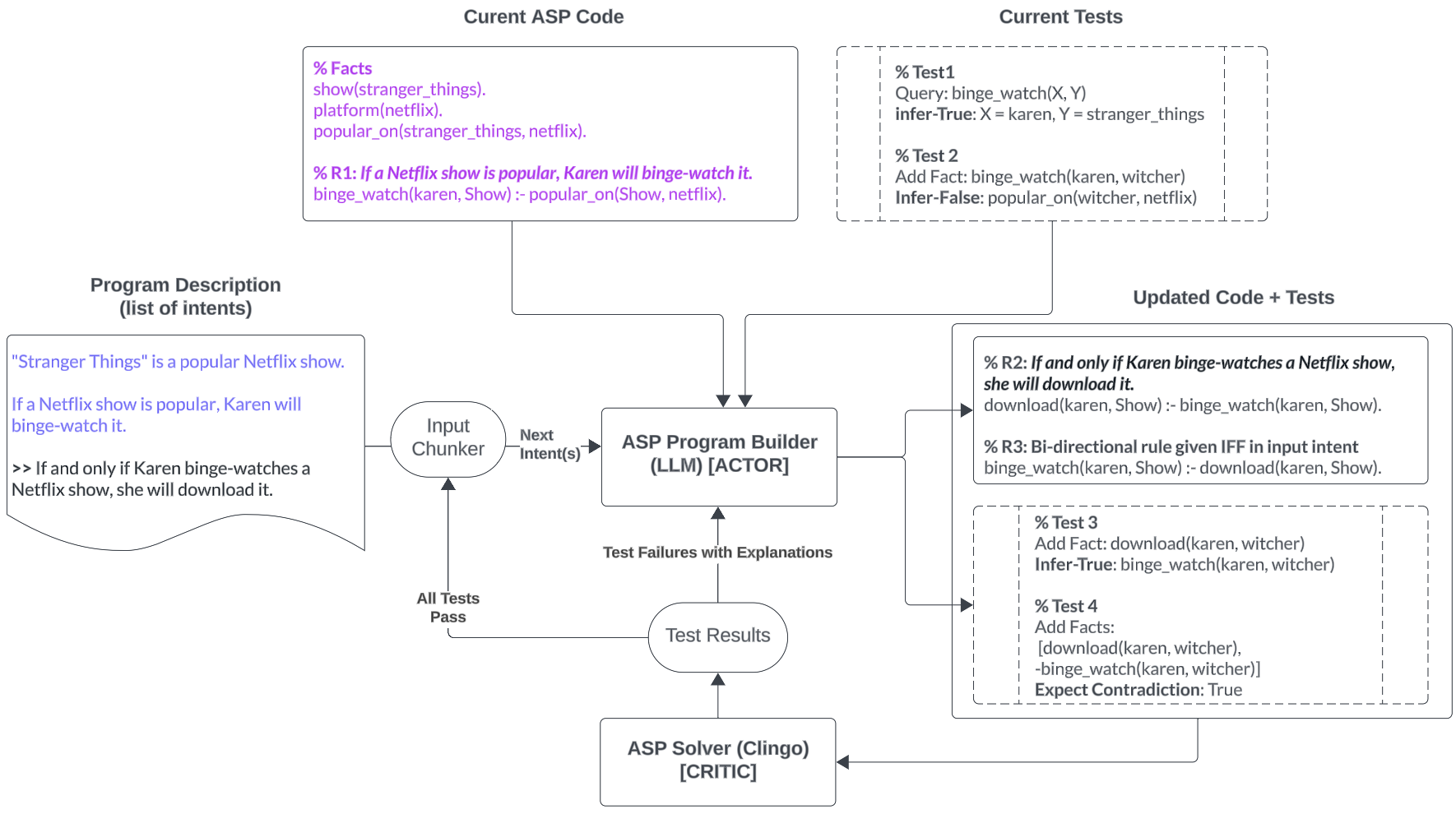}
    \captionsetup{font=small}
    \caption{\textbf{LLM-ARC Implementation based on Answer Set Programming (ASP)}: Given a problem description (a collection of natural language statements), the Actor (LLM) generates ASP code and tests in an iterative manner. At each step, the Actor takes as input the next segment (problem intents) to convert to ASP , along with the existing ASP code and tests generated so far, and outputs the updated code and tests based on the latest segment. The code is then run by the Critic (ASP Solver) and any test failures with explanations are fed back to the Actor. This self-correction loop runs till all tests-pass or max-iterations are reached. \textbf{The Actor is eventually trained on end-to-end dialog traces with the Critic feedback in this self-correction loop.}}
    \label{fig:llm-arc-design}
\end{figure*}


We use an LLM as the Actor and an Automated Reasoning engine as the Critic, and refer to this neuro-symbolic system as \textbf{LLM-ARC}. Figure \ref{fig:llm-arc-design} shows an implementation of LLM-ARC based on Answer Set Programming (ASP)\cite{asp}. In general, this design can apply to any LLM-code execution engine (replacing the automated reasoner with the corresponding code compiler/interpreter), though here we focus on declarative problem solving. 

Note that the system as designed above is not guaranteed to produce perfectly accurate results. This is because even if the code compiles without issues and all the generated tests pass, there is no guarantee that the test conditions correctly and completely capture the intended semantics, or that the tests pass for the right reason (e.g. the system could derive a required inference for a test using an incorrectly intended logical proof). We discuss this issue in Section \ref{sec:enhancements} and suggest a future enhancement using a separate Critic trained via human-feedback to evaluate the test criteria and reasoner results. 



To evaluate our LLM-ARC system, we run experiments on the FOLIO benchmark \cite{folio}. FOLIO is a human-annotated, logically complex and diverse dataset for reasoning in natural language. We use the latest version of FOLIO (v2) which contains 1001 training examples and 203 validation examples. The current state-of-the-art results on FOLIO is \textbf{78.9\%} achieved by LogicLM \cite{logiclm}. Using our LLM-ARC system we achieve a new state-of-the-art accuracy of \textbf{88.32\%}.

We compare several strong LLM-only baselines (using GPT4-Turbo as the LLM) with  various versions of the LLM-ARC system on the FOLIO data, and show that the Actor-Critic approach even in a few-shot setting (\textbf{only 8 examples} for the Actor) outperforms a fine-tuned LLM solution trained on all 1K examples.

We demonstrate that adding the test generation option to the Actor improves performance by \textbf{6.6\%} (compared to a version without test-gen); that running code and test generation in a self-correction loop with the Critic (where the Actor corrects mistakes based on the Critic feedback) further boosts performance by \textbf{~5\%}; and the best performing system is one where the Actor is trained on end-to-end self-correction dialog traces with Critic feedback (from the automated reasoner) on the training set. 

The contributions of this work are as follows:
\begin{itemize}[itemsep=1pt, topsep=1pt, partopsep=1pt, parsep=1pt]
\item We believe this is the first work to fold in test generation for declarative logic programs to improve code quality, and combine an LLM Actor for code-generation with a Reasoning Engine Critic for test evaluation and explanation, boosting overall system performance. (We refer to this hybrid architecture as LLM-ARC)
\item We specify guidelines for test-generation based on a logical analysis of the problem domain, and use a simple general schema for writing logic tests. We demonstrate the value-add of test generation, and the specific guidelines, via ablation experiments. All relevant LLM prompts are included in the Appendix.
\item In the presence of final ground truth labels for reasoning problems, we describe a fully automated procedure to train the Actor model (to write and rectify declarative code and tests) over end-to-end dialog traces of a self-correction loop using a reasoning engine Critic (to provide fine-grained explanatory feedback). This self-supervised version of the LLM-ARC system achieves a new SOTA of 88.32\% on the FOLIO benchmark.
\end{itemize}

\section{Related Work}
Given the remarkable performance of LLMs on automated code-generation, a large number of AI-driven ``co-pilot" tools and frameworks are being actively developed. There is also a growing interest in automatically generating test cases (both, unit tests and more complex integration tests) to validate code correctness \cite{testgen10298372, testgenalshahwan2024automated, testgenliu2024llmpowered}. However, to our knowledge, all the efforts have been focused on generating and testing procedural code. Our area of interest is symbolic code (logic programs) where tests are crucial to verify that the rules and constraints accurately captures the modelers intent. This is particularly useful for developers working with declarative systems who are not proficient in formal logic, due to the long-distance inter-dependencies across rules, vagaries of logic involving negations and contrapositives, and the need to explicitly encode commonsense knowledge. Moreover, we believe that test failures can be effectively leveraged to improve declarative code accuracy, due to the unique capability of a symbolic reasoning engines to provide detailed logical explanations (proofs) for the failures, a claim validated by our LLM-ARC system results.

More closely aligned to our work is neuro-symbolic systems such as LINC \cite{linc}
and LogicLM \cite{logiclm}. These systems combine an LLM with a formal reasoning engine (in LogicLM's case, a variety of solvers based on the underlying logic) and show impressive results on a several NL-reasoning benchmarks. However, neither system has the notion of generating semantic tests that need to be validated by the reasoner. LogicLM does have a self-refinement loop but it is only used for syntax errors in the generated logical representation, while LINC has no self-refinement or feedback from the solver. 

Additionally, the idea of training the logic program writer (Actor) over end-to-end interactions by incorporating feedback from a formal reasoning engine (Critic) is fundamentally novel to our work. Much of the ``agent based" or ``ReAct" systems that integrate tools with LLMs suffer from orchestration and control inefficiencies where individually efficient tools  are combined in a sub-optimal and brittle whole. We focus on integrated training that ensures the overall system is optimized.  

\section{Approach: Neuro-symbolic Actor-Critic Model}

As mentioned earlier, our LLM-ARC system is based on the Actor-Critic model, where we use an LLM as the Actor to generate declarative code with tests, and an Automated Reasoner as the Critic to execute the logic program, run the tests and provide detailed feedback with explanations to the Actor when there are test failures.

The system needs to be based on a logical formalism, and to tackle FOLIO, we chose \emph{Answer Set Programming (ASP)} as the underlying logic. ASP was selected because we found that it works best for developing enterprise applications \cite{chucarroll2024llms} and it has sufficient logical expressivity needed for most of the FOLIO problems.  

Figure \ref{fig:llm-arc-design} shows our LLM-ARC implementation based on ASP. We now describe details of the Actor and Critic.

\subsection{Actor: LLM Logic Program Writer}
For the LLM Actor, we chose GPT4-Turbo (\texttt{gpt-4-1106-preview}) since we found that it generated ASP code of reasonably high quality from NL instructions, even in a zero-shot setting. 

We use GPT4-Turbo in a few shot setting by specifying a handful of examples of translating FOLIO problems into ASP. To come up with the exemplar set, we did an automated analysis of the logical structure and expressivity of the NL statements in FOLIO.

\begin{figure*}[htb]
    \centering
    \includegraphics[width=\textwidth]{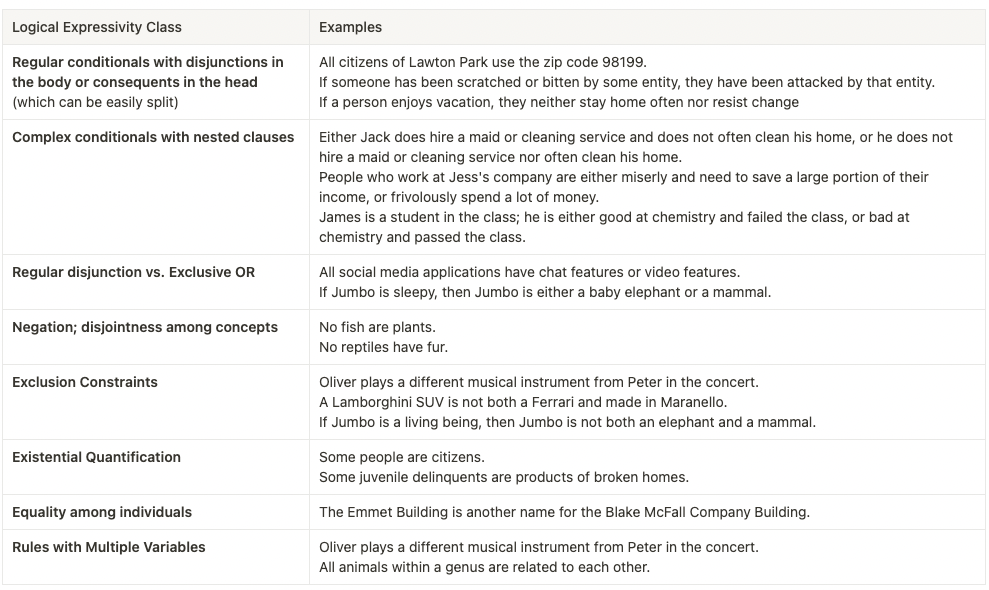}
    \captionsetup{font=small}
    \caption{\textbf{Logic Stratification of NL Statements in FOLIO}}
    \label{fig:logic-folio}
\end{figure*}

\subsubsection{Logic Stratification of FOLIO statements}
\label{sec:strata}
The idea is to use a powerful LLM (such as GPT4-Turbo) to automatically classify NL statements based on their logical structure, connectives/operators used, and overall composition (e.g. do they contain nested clauses). We came up with a general prompt (see Appendix) for logic stratification that applies to most formal logics (not just ASP) and ran it on a large random sample of FOLIO statements. We manually vetted the results and found that the logically stratified clusters (including their examples) found by the LLM were of very high quality overall. We acknowledge that this task may be easy for the FOLIO dataset where statements are written in a logic-heavy manner by design.

The net result of logic stratification over FOLIO data is shown in Figure \ref{fig:logic-folio}. The LLM found 8 logical classes of FOLIO statements. 

\begin{figure*}[htb]
    \centering
    \includegraphics[width=0.7\textwidth]{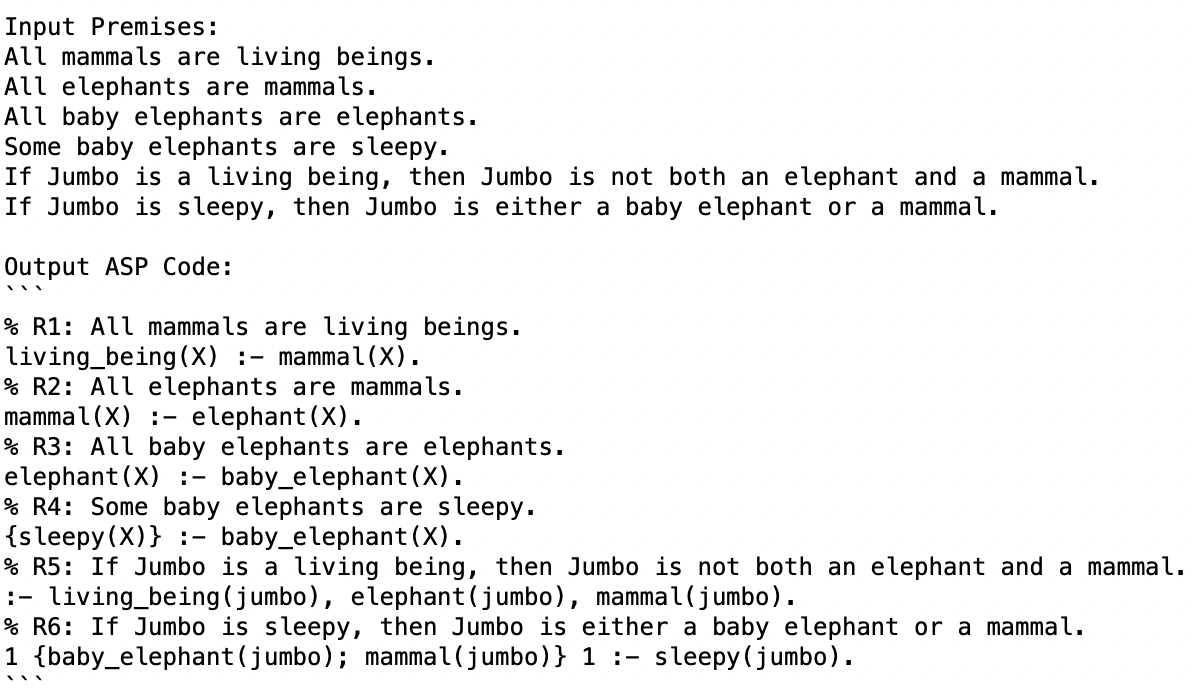}
    \captionsetup{font=small}
    \caption{\textbf{Training Example for NL to ASP used in the prompt} (In-Context Learning)}
    \label{fig:llm_writer_example}
\end{figure*}

We added one more category for cases where background knowledge (often common-sense relationships connecting predicates in the program) was missing in the input problem description. 

For example, consider when the Premises state: 

\emph{All employees who schedule a meeting with their customers will go to the company building today. Everyone who has lunch in the company building schedules meetings with their customers. No managers work remotely from home.}

In this case, the following common-sense rules are not explicitly mentioned in the premises:
\begin{itemize}[itemsep=1pt, topsep=1pt, partopsep=1pt, parsep=1pt]
\item \emph{Managers are employees.}
\item \emph{All employees who have lunch in the company building are in the company building.}
\end{itemize}

Information like this is often missing in the input because it is considered obvious, a problem noted by \cite{linc} as well. Here, we leverage the LLM's ability to fill in common-sense knowledge gaps, though we need to be careful about the LLM adding extraneous knowledge that confounds the modelers intent, and we address this using a combination of prompt-engineering and using tests to validate the semantics.

For the few shot setting, we added 8 examples to cover all the 8 main logic classes, adding one example per class (see Appendix). Several examples include common-sense relationships with instructions on how and when to add them. Note that since each example is a multi-line problem, a single example might cover more than one logic category. See an example in Figure \ref{fig:llm_writer_example}.
 
\subsubsection{Logic Test Generation}
We designed a simple general schema for specifying logic tests. Each test has optional facts that need to be added to the program to test the rules/constraints, and the test conditions are either one of the following:
\begin{itemize}[itemsep=1pt, topsep=1pt, partopsep=1pt, parsep=1pt]
\item \texttt{infer-True-All}: a set of propositions that must be inferred by the solver in \emph{all} solution sets of the logic program
\item \texttt{infer-True-Any}: a set of propositions that must be inferred by the solver in \emph{at least one} solution set of the logic program
\item \texttt{infer-False}: a set of propositions that must \emph{not be inferred} by the solver in any solution set of the logic program
\item \texttt{expect-Contradiction}: a boolean flag which represents whether we expect the program to be \emph{contradictory (unsatisfiable)} when the facts are added 
\end{itemize}

To improve test generation quality, we asked the LLM to add two additional fields for each test: \texttt{rules-referenced} - which points to specific rules in the program (all rules have an ID in the program; see the example in Figure \ref{fig:llm_writer_example}) that are exercised in the test; and \texttt{test explanation} - a rationale for the test describing how it validates the semantics of the referenced rules.

We then specified guidelines for writing tests in the prompt, based on different logical conditions in the input. Similar to the in-context examples chosen for ASP code generation, we mirrored the guidelines on the logic strata found in FOLIO statements, to ensure adequate coverage of the logical semantics. Examples of the guidelines are shown in Figure \ref{fig:test-guidelines}, with the full prompt attached in the Appendix.

\begin{figure*}[htb]
    \centering
    \includegraphics[width=0.95\textwidth]{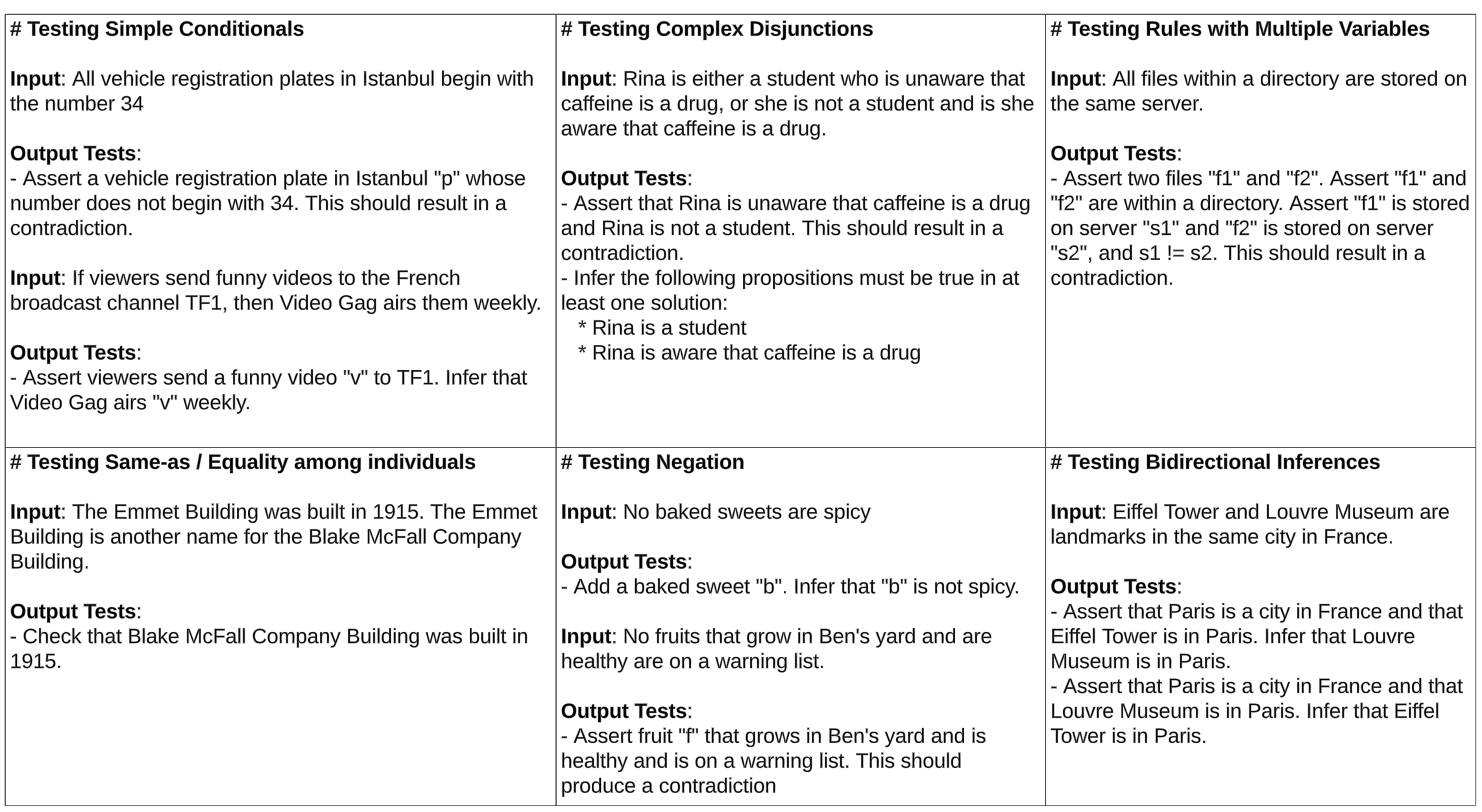}
    \captionsetup{font=small}
    \caption{\textbf{Test Guidelines for the various logic classes with examples}}
    \label{fig:test-guidelines}
\end{figure*}

\subsubsection{Error Correction}
Finally, we include instructions in the Writer prompt for correcting errors reported by the Critic. There are two kinds of errors: \emph{syntax/compilation} errors, and \emph{semantic errors} when there are test failures. The prompt contains strategies to resolve both classes of errors, and utilizes 1 example each. Furthermore, we specify a detailed pseudo-code for fixing the semantic errors (failing tests) since this is the more challenging case. The instructions walk through how the explanation from the Critic can be used to identify whether the test inputs, validation criteria, or specific parts of the ASP program (e.g., the commonsense knowledge section) need to be altered.

\subsection{Critic: Logical Reasoner}
We use the Clingo ASP Solver \cite{clingo} as the Critic since it is highly performant and freely available under the MIT License. Clingo also has useful compilation error messages, which point to specific lines in the program with errors. This information is fed back to the Actor in the self-correction loop.

\subsubsection{Query Evaluation}
We came up with a simple logical grammar to interpret the Conclusion statements in the FOLIO problem as structured queries, which could then be evaluated more accurately using the Solver. 

For example, consider the following Conclusion: \emph{``If the Red Star is a supernova or observed for its brightness, then the Red Star is neither a planet nor is its orbit stable."}

This is the corresponding target interpretation:
\begin{enumerate}[itemsep=1pt, topsep=1pt, partopsep=1pt, parsep=1pt]
\small
    \item \texttt{ATOM(supernova(red\_star))}
    \item \texttt{ATOM(observed\_for\_brightness(red\_star))}
    \item \texttt{OR(1, 2)}
    \item \texttt{ATOM(-planet(red\_star))}
    \item \texttt{ATOM(-orbit\_stable(red\_star))}
    \item \texttt{AND(4, 5)}
    \item \texttt{IF-THEN(3, 6)}
\end{enumerate}

As shown, we use standard logical operators such as \texttt{And, Or, Not, XOR, IF-THEN} (for implications) and use \texttt{ATOM} to denote the base atomic propositions. Any logical structure can be composed bottom-up in a modular manner. The advantage of representing queries (conclusions) using this schema is more flexibility in query evaluation, especially when dealing with the particularities of the ASP formalism (e.g. ASP does not have clean support for existential quantification). 

\subsubsection{Explanation Generation}
A feature that we added to the Solver is its ability to generate explanations for query entailments. There has been some work in this area \cite{asp-explanation} though we developed our own simple algorithm based on \emph{proof-by-refutation}. The idea is to check query entailment by adding the negation of the query to the program and checking for a contradiction. If a contradiction is found, we can infer that the query is entailed. In this case, we can find an explanation for the entailment by obtaining the minimal set of rules in the ASP program that result in the contradiction. This is a popular technique for explanation generation used in FOL and description logic systems \cite{dl-explanations}. 

\section{Experiments on FOLIO}
We conducted various experiments using the FOLIO benchmark, which consists of 1001 training examples and 204 validation examples. 

Each FOLIO problem consists of a set of \emph{Premises} (NL statements) and a \emph{Conclusion} (also a NL statement). The task is to determine whether the Conclusion is \texttt{True, False} or \texttt{Uncertain} given the Premises. The FOLIO dataset also includes First-Order-Logic (FOL) translations for each of the Premises and the Conclusion.

We evaluated FOLIO use the following systems (the first four systems below are \emph{LLM-only} baseline systems)

\begin{enumerate}[itemsep=1pt, topsep=1pt, partopsep=1pt, parsep=1pt]
\item \textbf{GPT-3.5-ZS and GPT4-T-ZS}: Zero-shot versions of GPT-3.5 and GPT4-Turbo
\item \textbf{GPT4-T-CoT}: GPT4-Turbo with a Chain-of-Thought prompt where we instruct the model to label the premises, and then carefully evaluate the conclusion using step-wise reasoning and referencing the premises along the way.
\item \textbf{GPT4-FT-NL}: GPT4 fine-tuned on the NL problem descriptions in the entire FOLIO training data of 1001 examples
\item \textbf{GPT4-FT-FOL}: GPT4 fine-tuned to go from NL problem description to the corresponding First Order Logic (FOL) versions (annotated in the FOLIO training data), and then to the prediction. The idea is to check whether using the precise FOL translations as an intermediate step helps the model produce more accurate results.
\item \textbf{LLM-ARC-8-shot}: The LLM-ARC system with 8 in-context learning examples, where the LLM Actor only does code generation (no Tests). The LLM used was GPT4-Turbo
\item \textbf{LLM-ARC-8-shot-TestGen}: The above system with the enhancement that the Actor also generates Tests for the code
\item \textbf{LLM-ARC-20-shot}: The LLM-ARC system (again using GPT4-Turbo as the Actor) with 20 in-context learning examples, and no test generation. We added another 14 examples to cover the 8 logic classes described in Section \ref{sec:strata}.
\item \textbf{LLM-ARC-20-shot-TestGen}: The above system with the enhancement that the Actor also generates Tests for the code
\item \textbf{LLM-ARC-Trained}: Trained version of the LLM Actor (GPT4, not Turbo) on end-to-end dialog traces with the Reasoner Critic, in a self-correction loop over the entire training data. The actor is trained to generate both ASP Code and Tests. Details of how this was done are provided in the next subsection.
\end{enumerate}

All the LLM-ARC systems are run in a self-correction loop with upto 4 iterations.

\subsection{Training the Actor with Critic Feedback Dialog-Traces}
We ran the un-trained 8-shot version of the LLM-ARC system (with the TestGen capability) on the entire training set and collected dialog trace data on the correctly predicted examples. We used this dialog data to fine-tune a separate Actor model based on GPT4\footnote{Currently, OpenAI does not provide an option to fine-tune GPT4-Turbo.}. Since the context window of GPT4 is only 8K tokens, we had to limit the dialog traces to fit it into the window. We achieved this by using only the \emph{last rectification step} of the trace -- e.g. if there was a compiler error reported by the Critic that was fixed by the Actor in the next iteration, we would train on a trace that starts with the prior incorrect version from the Actor, followed by the Critic feedback, and then the corrected version with the compilation issues fixed. The same applies to test failures, where we started the dialog trace with a version just prior to all the tests being passed, and included the intermediate Critic feedback before the corrected version. 

Additionally, we included a two-step ``short-cut" dialog trace that went from the input problem directly to the ASP code and tests, when all the tests passed along with the correct ground truth prediction. The idea behind this is to enable the Actor to learn how to produce code and tests of high quality (that compile, pass tests and entail the query correctly) in a direct manner. 

To summarize, the data used to fine-tune the GPT4 Actor had the following 3 kinds of dialog traces:
\begin{enumerate}[itemsep=1pt, topsep=1pt, partopsep=1pt, parsep=1pt]
\item NL description $\rightarrow$ ASP code with compilation issues $\rightarrow$ Critic Feedback on compiler errors $\rightarrow$ ASP code that compiles
\item NL description $\rightarrow$ ASP code that compiles with test failures $\rightarrow$ Critic feedback on failures with explanations $\rightarrow$ ASP code with all tests passing
\item NL description $\rightarrow$ ASP code that compiles with all tests passing
\end{enumerate}
The traces were collected whenever the final system prediction on the ground truth label was correct. The total number of dialog traces (each trace corresponds to a single training instance) collected on the entire training set was 918.

Finally, to keep the LLM prompt in the fine-tuned training data as concise as possible, we did not include any examples in the trained Actor prompt. 

\begin{figure*}
    \centering
    \includegraphics[width=1.1\textwidth]{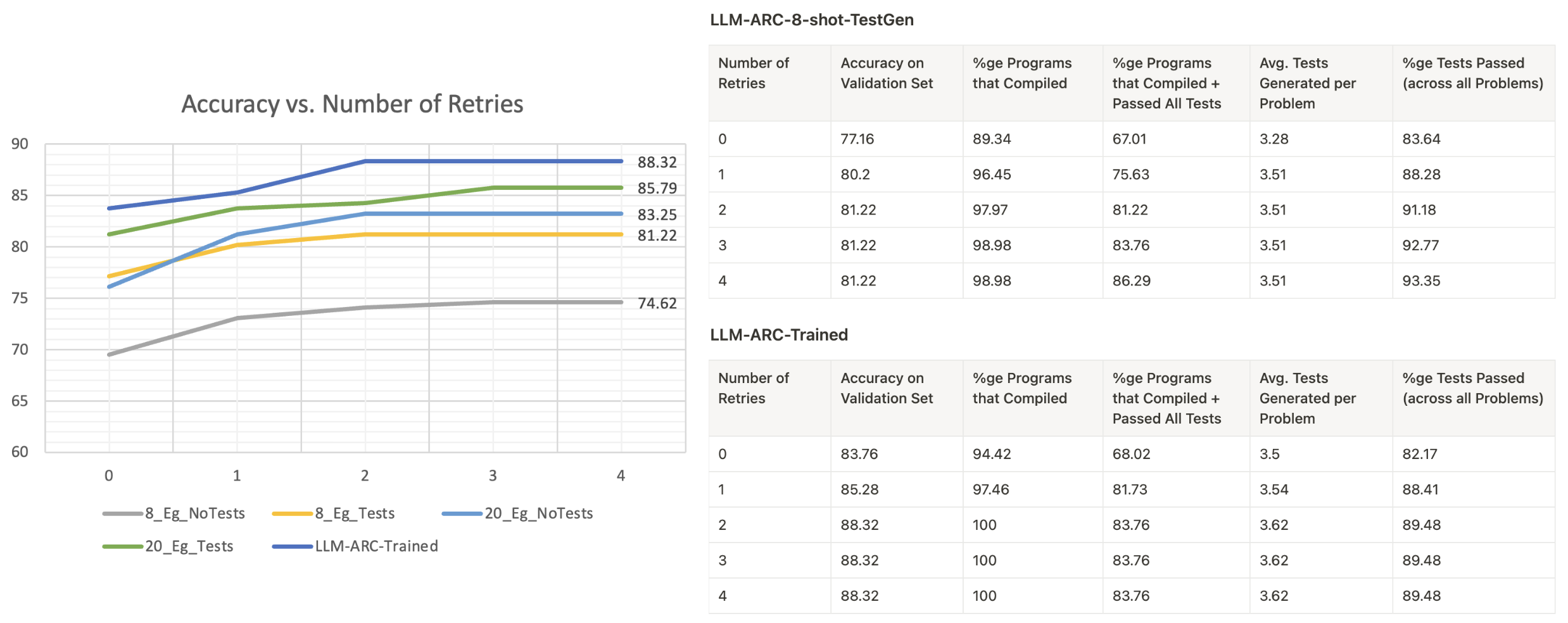}
    \captionsetup{font=small}
    \caption{\textbf{Impact of Iterative Self-Correction} Over multiple iterations, we see overall LLM-ARC system accuracy go up, as more ASP programs fully compile and pass more tests. The chart on the left shows system accuracy over multiple iterations for the various LLM-ARC system variants. The two tables on the right show additional statistics around generated tests, code compilation and test passing for the LLM-8-shot and Trained version.} 
    \label{fig:retries}
\end{figure*}

\subsection{System Results}

\begin{minipage}{\columnwidth}
    \centering
    \begin{tabularx}{\columnwidth}{|p{5.5cm}|p{1.75cm}|} \hline
        \textbf{System} & \textbf{Accuracy} \\
        \hline
        GPT3.5-ZS & 66.9\% \\ \hline
        GPT4-T-ZS & 67\% \\ \hline
        GPT4-T-CoT & 74.1\%  \\ \hline
        GPT4-FT-NL & 80.7\% \\ \hline
        GPT4-FT-FOL & 78.17\% \\ \hline
        LogicLM (Prior SOTA) & 78.9\% \\ \hline
        LLM-ARC-8-shot & 74.62\% \\ \hline
        LLM-ARC-8-shot-TestGen & 81.22\% \\ \hline
        LLM-ARC-20-shot & 83.25\% \\ \hline
        LLM-ARC-20-shot-TestGen & 85.79\% \\ \hline
        LLM-ARC-Trained & \textbf{88.32\%} \\ \hline
    \end{tabularx}
    \captionof{table}{\textbf{Overall Accuracy on FOLIO.} All LLM-ARC systems were run in a self-correction loop with upto 4 iterations} 
    \label{table:main-results}
\end{minipage}
\\

The accuracy scores for all the systems are shown in Table, \ref{table:main-results}, along with the prior known SOTA\footnote{This result was reported on FOLIO v1. We are in the process of replicating their system's results on the latest v2 version.}.

The best performing LLM-only baseline solution is \texttt{GPT4-FT-NL} which was trained on all the 1K examples in the training data. Interestingly, using the FOL annotations as intermediate representations in a chain-of-thought variant did not help results, a point worth investigating in the future.

Regarding the LLM-ARC systems, we observe a clear benefit of adding Test Generation. Both the LLM-ARC few shot variants (i.e. 8 and 20 example variants) perform better with TestGen added, with the 8-example variant seeing a huge boost of +6.6\%. Of note, the \texttt{LLM-ARC-8-shot-TestGen} version outperforms the best LLM-only solution even though the latter was fine-tuned on the entire 1K example training set.

Our best performing system is the LLM-ARC version that was trained in a self-supervised manner on end-to-end dialog traces with the Critic feedback, and achieved 88.32\% accuracy, 10 points higher than the prior known SOTA.

\subsection{Ablation Studies}
We conducted a few ablation studies to measure the impact of features in the LLM-ARC system.

\subsubsection{Impact of Iterative Self-Correction}
To answer the question ``How much do retries help?", we plot the accuracy curves for the LLM-ARC variants over multiple iterations of the self-correction loop. The results are shown in Figure \ref{fig:retries}. We see that over multiple iterations, the performance does go up by between 4-5\% for the two LLM-ARC systems shown above, when comparing the results after zero and max-retries, as the code compilation issues and test failures are fixed by the Actor based on the Critic feedback (notice how the numbers in columns 3 and 6 in the Tables in Figure \ref{fig:retries} go up over iterations along with the overall accuracy in column 2). However, the final accuracy asymptotes after two retries. More details on this are in the Error Analysis section. 

\subsubsection{Impact of TestGen Guidelines}
To measure the impact of adding Test Generation guidelines (Figure \ref{fig:test-guidelines}), we ran an ablation by dropping this section from the prompt and letting the LLM determine how to generate tests on its own instead. The results for this ablation are shown in the table below for the two LLM-ARC few-shot systems, and indicate a big drop in performance, clearly demonstrating its value-add. 
\\
\\
\begin{minipage}{\columnwidth}
    \centering
    \begin{tabularx}{\columnwidth}{|p{4.5cm}|p{2.75cm}|} \hline
        \textbf{System} & \textbf{Accuracy} \\
        \hline
        LLM-ARC-8-TestGen & 78.7\% (-2.52\%) \\ \hline
        LLM-ARC-20-TestGen & 80.2\% (-5.59\%) \\ \hline
    \end{tabularx}
    \captionof{table}{\textbf{Dropping Test-Gen Guidelines}}
    \label{table:ablate-testgg}
\end{minipage}
\\


\section{Error Analysis and Discussion}
We analyzed errors from the best performing system (\texttt{LLM-ARC Trained}) and found that they broadly fell in three categories (excluding minor cases of query interpretation failures and modeling mistakes like representing an XOR as a regular disjunction)
\begin{enumerate}[itemsep=1pt, topsep=1pt, partopsep=1pt, parsep=1pt]
\item \textbf{Existential quantification}: ASP does not have natural support for existential quantification. For example, it is not possible to accurately model the following statement: \emph{``One six-way tie was on the leaderboard, and one person in the six-way tie was from Belgium."} that posits the existence of two unnamed individuals, which can potentially be unified with other named individuals in the program. This is certainly possible to do in other logics such as FOL (which the FOLIO dataset was annotated with) but is a limitation of our chosen formalism.
\item \textbf{Rules with Multiple Variables}: Among the various logic classes in FOLIO identified in Figure \ref{fig:logic-folio}, the one class that the Actor had difficulty in modeling was rules with multiple variables. This includes statements like \emph{``All languages within a language family are related to each other."}, where the rule involves two variables for distinct languages. We believe a reason for this is that there are very few examples of this class in the training set ($<5\%$). A potential solution is to up-weight (or up-sample) examples from this class during training, or simply give more examples of this class in the few-shot prompt. 
\item \textbf{Conflating types and instances}: These are cases where certain entities are linguistically used as both types and individuals in the input problem. For example, consider the example below from FOLIO where we have marked an entity that looks like both a class and an individual in different statements with a ``*".

\begin{verbatim}
    Plungers suck.
    *Vacuums* suck.
    Vampires suck.
    Space is a *vacuum*.
    A duster is a household 
    appliance that doesn't suck.    
\end{verbatim}

Similar to the previous error class, there are very few examples of this behavior in the training set. Moreover, these are particularly hard modeling problems from a logic standpoint in ASP, which does not support ``punning" (as say the description logic OWL2\footnote{https://www.w3.org/2007/OWL/wiki/Punning}), and hence requires additional machinery at modeling and query evaluation time to correctly interpret terms as either classes or individuals based on how they are used. 
\end{enumerate}

Given the high performance of the \texttt{LLM-ARC-Trained} system, it is unsurprising that the remaining headroom is for the challenging or sparse cases. 

Finally, we looked into why the performance was asymptotic after a few retries of the self-correction loop. In roughly a third of the problem failure categories (i.e. final prediction was incorrect) and where all tests did not pass, we found that the Actor made no alterations to the program code or the tests from one iteration to the next. This is a weakness exposed by our current design where we do not enforce that some alteration must be made by the Actor, and instead expect the LLM to follow the instructions in the prompt, which it clearly does not always do. We are considering an alternate design using function-calling with constraints where we can enforce that one or more of the program code, query interpretation and failing test criteria must be modified in the presence of a failing test. We also empirically observed that the Actor would rarely change the query interpretation across multiple iterations and found that this was a miss in our instructions which primarily focused on altering the program code and tests.

\subsection{Potential Enhancements}
\label{sec:enhancements}

There are several potential enhancements to the LLM-ARC implementation which we leave for future investigation:

\textbf{Sophisticated Input Chunking}: Since FOLIO problems are relatively small ($<$ 10 statements), we pass the entire problem to the LLM Actor in one shot, without doing any chunking. In the future, for real world applications that involve translating large volumes of business logic text into a formal program, the input would have to be chunked. The appropriate chunking level would depend on the quality of the code generated, and would have to be empirically determined based on the LLM's output quality given a certain context window size (and if the Actor is trained, the chunking size used in the training data).

\textbf{Enhancing Critic Explanations}: The current explanation generated from Clingo using our proof-by-refutation algorithm does not include grounded statements. A more informative explanation would come from grounding relevant rules that lead to the entailment, as described in work \cite{asp-explanation}. Moreover, we could use another LLM to translate the grounded rule-based proof back into natural language to produce a more fluent explanation, which should presumably be more interpretable for the LLM Actor. This hypothesis needs to be empirically validated.

\textbf{Training a separate Critic}: As mentioned earlier, 
in the current design, there is no guarantee that the test conditions correctly and completely capture the intended semantics, or that the tests pass for the right reason. One way to mitigate this issue is to have a \emph{separate} Critic that evaluates the reasoner's results and provides feedback on the test criteria and proof step correctness. Indeed, our original system design started off with a Critic distinct from the reasoner, which was to be trained with human-feedback on the tests results and explanations provided by the reasoner (since those need to be manually assessed). We did not go down this path in the end, since we found that using the automated reasoner as the Critic directly, and training the Actor in a self-supervised training loop, produced a big boost in performance. We still believe training a separate critic has the potential to further increase accuracy and reliability of the entire system.


\section{Conclusion}
There is growing recognition in the AI community that LLM-only solutions do not meet the standard for production applications that require a high degree of accuracy, consistency and explicability. More specifically, current state-of-the-art LLMs are known to struggle for problems involving precise logical reasoning, planning and constraint solving. As a result, we have seen a rise in the development of Neuro-Symbolic systems, where the reasoning is offloaded to a symbolic solver, and the LLM is used at the interface layer to map between unstructured data (text) and structured logical representations. Unlike standard tools or simple APIs, integration between an LLM and a symbolic reasoner can be fairly sophisticated as the reasoning engine has its own world model and decision procedures (arguably, one might even conceive and design the system such that the reasoner is the brain of the system and the LLM is the tool for interpreting and translating data). 

In such declarative systems, we firmly believe that tests are needed to check for semantic correctness of the logic program (a much harder challenge than ensuring syntactic correctness), and that the reasoner by way of providing detailed feedback on test failures to the program writer can help it improve in a self-correction loop. This intuition led us to the design the \texttt{LLM-ARC} system presented in this paper, which is based on the Actor-Critic model and uses the LLM as the Actor and an Automated Reasoning engine as the Critic. We empirically validate the system on the FOLIO benchmark, and show that not only can such a system achieve higher performance than an LLM-only solution in a few-shot setting, but that we can devise a fully automated self-supervised loop to train the Actor with Critic feedback to  boost performance significantly. Lastly, the ability of this system to provide detailed logical explanations for its answers means that a human-in-the-loop can verify its results in production applications. 

\bibliographystyle{plain}
\bibliography{bibliography}

\end{multicols}

\section{Appendix}

\subsection{Prompt for Logic Stratification}
You are an expert logician. You are given a set of natural language statements that express various logical conditions, rules and constraints. Your task is to stratify (cluster) the statements based on their logical structure, connectives (or operators used) and complexity (e.g., nested clauses etc.). 

Output a list of clusters where each cluster contains a collection of statements that have a similar logical structure and connectives used. Copy up to 5 canonical problem statements in each cluster. 

Be as fine-grained as possible when coming up with clusters and come up with an exhaustive set of clusters that cover all the diversity in the input statements.

\subsection{8-Examples (one per logic class) used in LLM Actor}
\begin{verbatim}
#### Example 1
Input:
Either Jack does hire a maid or cleaning service and does not often clean his home, 
or he does not hire a maid or cleaning service nor often clean his home.
Query: Jack hires a maid or cleaning service.

Output Model:
% R1: Jack has two options
1 {jack_option1; jack_option2} 1.
% R2: Jack option 1 is: Jack hires a maid or cleaning service 
% and does not often clean his home
1 {hires_maid_or_service(jack); -often_clean(jack)} 1 :- jack_option1.
% R3: Jack option 2 is: Jack does not hire a maid or cleaning service
% and does not often clean his home
-hires_maid_or_service(jack) :- jack_option2.
-often_clean(jack) :- jack_option2.

Output Query:
1. ATOM(hires_maid_or_service(jack))

#### Example 2
Input:
Some people are dependent on coffee. Rina is either a student who 
is dependent on caffeine, or she is neither.
Query: Some people are students.

Output Model:
% R1: Some people are dependent on coffee
{dependent_on_caffeine(X)} :- person(X).
% R2: Rina is either a student and dependent on caffeine, or she is neither.
1 {rina_student_dependent; rina_neither} 1.
% R3: Add rules to split the conjunction "rina_student_dependent"
student(rina) :- rina_student_dependent.
dependent_on_caffeine(rina) :- rina_student_dependent.
% R4: Add rules to split the conjunction "rina_neither"
-student(rina) :- rina_neither.
-dependent_on_caffeine(rina) :- rina_neither.
% ISA Rules
% R5: All students are people
person(X) :- student(X).

Output Query:
1. ATOM(person(X))
2. ATOM(student(X))
3. SOME(1,2)

#### Example 3
Input:
No fish are plants. Trees are plants. Fir is a tree.
Query: Fir is not a fish.

Output Model:
% R1: No fish are plants.
-plant(X) :- fish(X).
% R2: Trees are plants.
plant(X) :- tree(X).
% R3: Fir is a tree.
tree(fir).
% ISA Rules
% R4: "No fish are plants" also means no plants are fish. 
% This rule is needed for the query.
-fish(X) :- plant(X).

Output Query:
1. ATOM(-fish(fir))

#### Example 4
Input:
Tom is a Russian choral conductor who specialized in French Renaissance.
Query: No choral conductor specialized in the performance of Renaissance

Output Model:
% R1: Tom is a Russian choral conductor
russian_choral_conductor(tom).
% R2: Tom specialized in French Renaissance
% specialized_french_renaissance(tom).
% ISA Rules
% R3: A Russian choral conductor is also a choral conductor
choral_conductor(X) :- russian_choral_conductor(X).
% R4: A specialized french renaissance is a french renaissance
french_renaissance(X) :- specialized_french_renaissance(X).

Output Query:
1. ATOM(choral_conductor(X))
2. ATOM(specialized_renaissance(X))
3. AND(1, 2)
4. NOT(3)

#### Example 5
Input:
The Blake McFall Company Building is a building added to the National 
Register of Historic Places in 1990.
The Emmet Building is a five-story building in Portland, Oregon.
The Emmet Building was built in 1915.
The Emmet Building is another name for the Blake McFall Company Building.
John works at the Emmet Building.
Query: John started his current job in 1990

Output Model:
% R1: The Blake McFall Company Building is a building added to the
% National Register of Historic Places in 1990.
historic_building(blake_mcfall_company_building, 1990).
% R2: The Emmet Building is a five-story building in Portland, Oregon.
five_story_building(emmet_building).
location(emmet_building, portland_oregon).
% R3: The Emmet Building was built in 1915.
built_year(emmet_building, 1915).
% R4: The Emmet Building is another name for the Blake McFall Company Building.
same_building(emmet_building, blake_mcfall_company_building).
% R5: John works at the Emmet Building.
works_at(john, emmet_building).
% R6: Same building is symmetric
same_building(X, Y) :- same_building(Y, X).
% R7: same_building needs to apply to other predicates involving buildings
historic_building(X, Z) :- same_building(X, Y), historic_building(Y, Z).
five_story_building(X) :- same_building(X, Y), five_story_building(Y).
location(X, Z) :- same_building(X, Y), location(Y, Z).
built_year(X, Z) :- same_building(X, Y), built_year(Y, Z).
works_at(Z, X) :- same_building(X, Y), works_at(Z, Y).

Output Query:
1. ATOM(started_job(john, 1990))

#### Example 6
Input:
A werewolf is a human that can turn into a wolf.
A werewolf has been scratched or bitten by another werewolf.
If someone has been scratched or bitten by some entity, 
they have been attacked by that entity.
Query: All humans are werewolves.

Output Model:
% R1: A werewolf is a human that can turn into a wolf.
werewolf(X) :- human(X), can_turn_into_wolf(X).
% R2: A werewolf has been scratched or bitten by another werewolf.
scratched_or_bitten_by_werewolf(X) :- werewolf(X).
% R3: If someone has been scratched or bitten by some entity, 
% they have been attacked by that entity.
attacked_by(Y, X) :- scratched_or_bitten_by(Y, X).
% R4: A werewolf has been scratched or bitten by another werewolf implies 
% the werewolf has been attacked by another werewolf.
attacked_by_werewolf(X) :- scratched_or_bitten_by_werewolf(X).
% ISA Rules: Need to connect generalized predicates to specialized versions
% R5: If someone is "attacked_by" a werewolf, then they are "attacked_by_werewolf"
attacked_by_werewolf(X) :- attacked_by(Y, X), werewolf(X).
% R6: If someone is "scratched_or_bitten_by" a werewolf, then 
% they are "scratched_or_bitten_by_werewolf"
scratched_or_bitten_by_werewolf(X) :- scratched_or_bitten(Y, X), werewolf(X).

Output Query:
1. ATOM(human(Human))
2. ATOM(werewolf(Human))
3. ALL(1, 2)

#### Example 7
Input:
All people in this tech company who are consistent and enjoy sticking to their regular 
routines do not like surprises.
People in this tech company who wear the same flannel shirts every day are  consistent 
and enjoy sticking to their regular routines.
People in this tech company who do not like shopping for clothes wear the same flannel 
shirts every day.
Old people living in stable homes do not like surprises.
People in this tech company who have very high energy and are impulsive like surprises.
Mike works in this tech company.
If Mike is not a person who wears the same flannel shirts every day, has very high energy, 
and is impulsive, then Mike either is very consistent and enjoys sticking to his
regular routines or does not like surprises.
Query: If Mike wears the same flannel shirts every day or does not 
like shopping for clothes, then Mike is neither an old person living 
in a stable home nor does he like shopping for clothes.

Output Model:
% R1: All people in this tech company who are consistent and 
% enjoy sticking to their regular routines do not like surprises.
-likes_surprises(X) :- consistent(X), enjoys_regular_routines(X),
  works_in_tech_company(X).
% R2: People in this tech company who wear the same flannel shirts 
% every day are consistent and enjoy sticking to their regular routines.
consistent(X) :- wears_flannel_shirts_every_day(X), works_in_tech_company(X).
enjoys_regular_routines(X) :- wears_flannel_shirts_every_day(X),
  works_in_tech_company(X).
% R3: People in this tech company who do not like shopping for clothes 
% wear the same flannel shirts every day.
wears_flannel_shirts_every_day(X) :- does_not_like_shopping_for_clothes(X), 
works_in_tech_company(X).
% R4: Old people living in stable homes do not like surprises.
-likes_surprises(X) :- old_person(X), lives_in_stable_home(X).
% R5: People in this tech company who have very high energy
% and are impulsive like surprises.
likes_surprises(X) :- very_high_energy(X), impulsive(X), 
  works_in_tech_company(X).
% R6: Mike works in this tech company.
works_in_tech_company(mike).
% R7: If Mike is not a person who wears the same flannel shirts every day, 
% has very high energy, and is impulsive, then Mike either is very 
% consistent and enjoys sticking to his regular routines or does not like surprises.
1 {consistent_and_enjoys_regular_routines(mike); -likes_surprises(mike)} :- 
  -wears_flannel_shirts_every_day(mike), very_high_energy(mike), impulsive(mike).
consistent(mike) :- consistent_and_enjoys_regular_routines(mike).
enjoys_regular_routines(mike) :- consistent_and_enjoys_regular_routines(mike).

Output Query:
1. ATOM(wears_flannel_shirts_every_day(mike))
2. ATOM(does_not_like_shopping_for_clothes(mike))
3. OR(1, 2)
4. ATOM(-old_person(mike))
5. ATOM(-does_not_like_shopping_for_clothes(mike))
6. AND(4, 5)
7. IF-THEN(3, 6)

#### Example 8
Input:
All mammals are living beings.
All elephants are mammals.
All baby elephants are elephants.
Some baby elephants are sleepy.
If Jumbo is a living being, then Jumbo is not both an elephant and a mammal.
If Jumbo is sleepy, then Jumbo is either a baby elephant or a mammal.
Query: Jumbo is sleepy.

Output Model:
% R1: All mammals are living beings.
living_being(X) :- mammal(X).
% R2: All elephants are mammals.
mammal(X) :- elephant(X).
% R3: All baby elephants are elephants.
elephant(X) :- baby_elephant(X).
% R4: Some baby elephants are sleepy.
{sleepy(X)} :- baby_elephant(X).
% R5: If Jumbo is a living being, then Jumbo is not both an elephant and a mammal.
:- living_being(jumbo), elephant(jumbo), mammal(jumbo).
% R6: If Jumbo is sleepy, then Jumbo is either a baby elephant or a mammal.
1 {baby_elephant(jumbo); mammal(jumbo)} 1 :- sleepy(jumbo).

Output Query:
1. ATOM(sleepy(jumbo))
\end{verbatim}

\subsection{Full Prompt for LLM Actor}
\begin{figure*}[htb]
    \centering
    \includegraphics[width=0.75\textwidth]{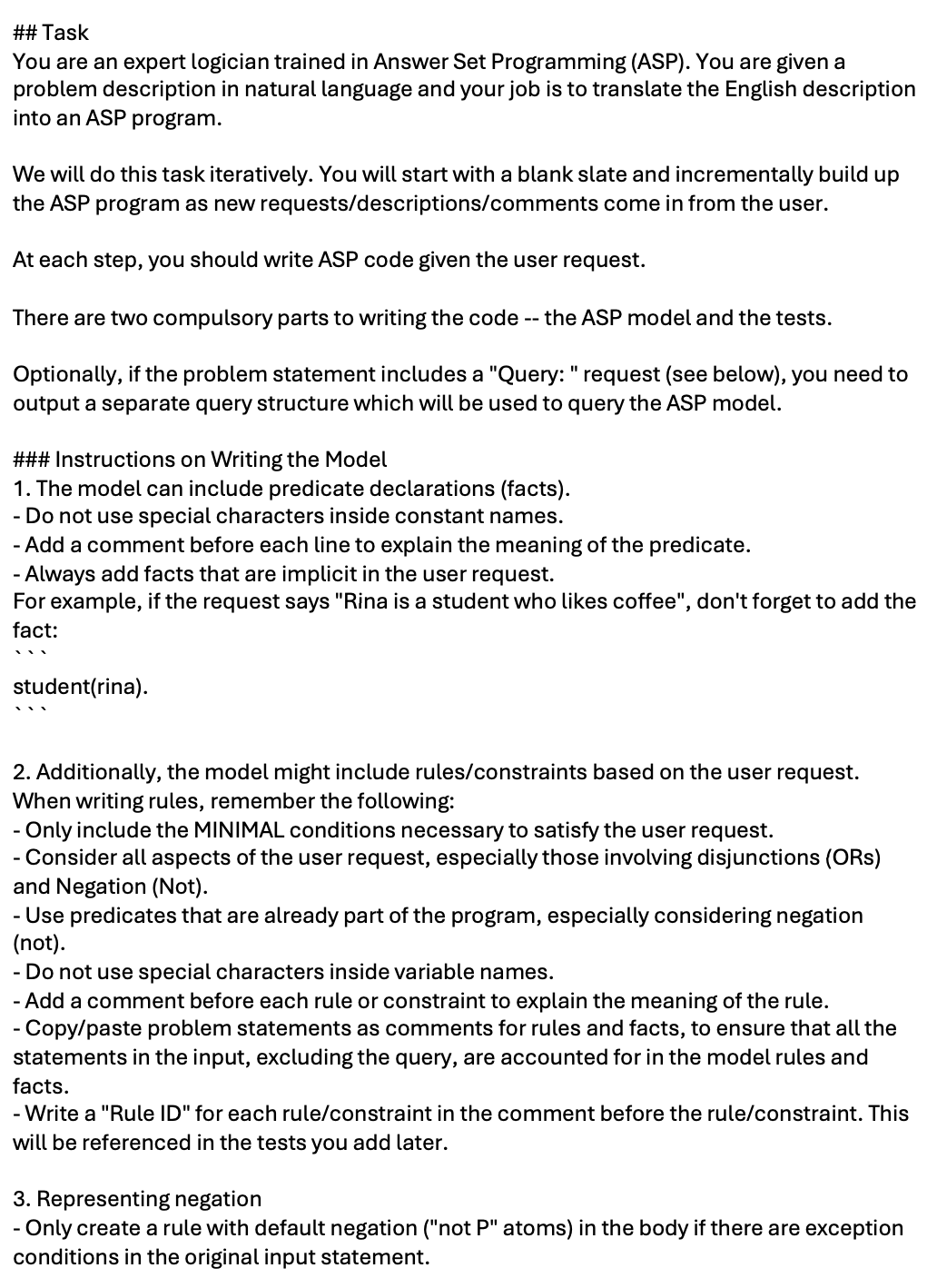}
\end{figure*}
\begin{figure*}
    \centering
    \includegraphics[width=0.75\textwidth]{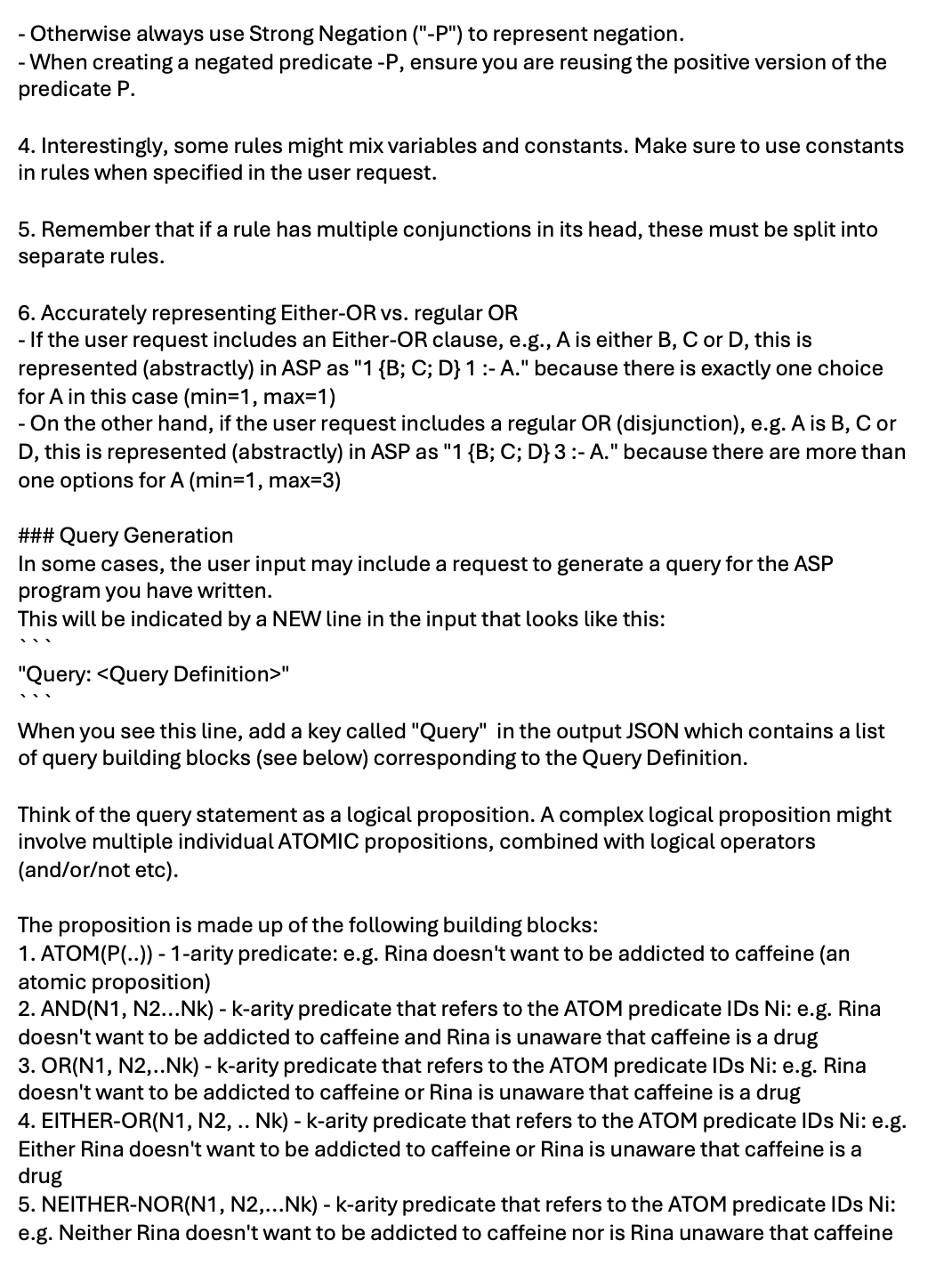}
\end{figure*}
\begin{figure*}
    \centering
    \includegraphics[width=0.75\textwidth]{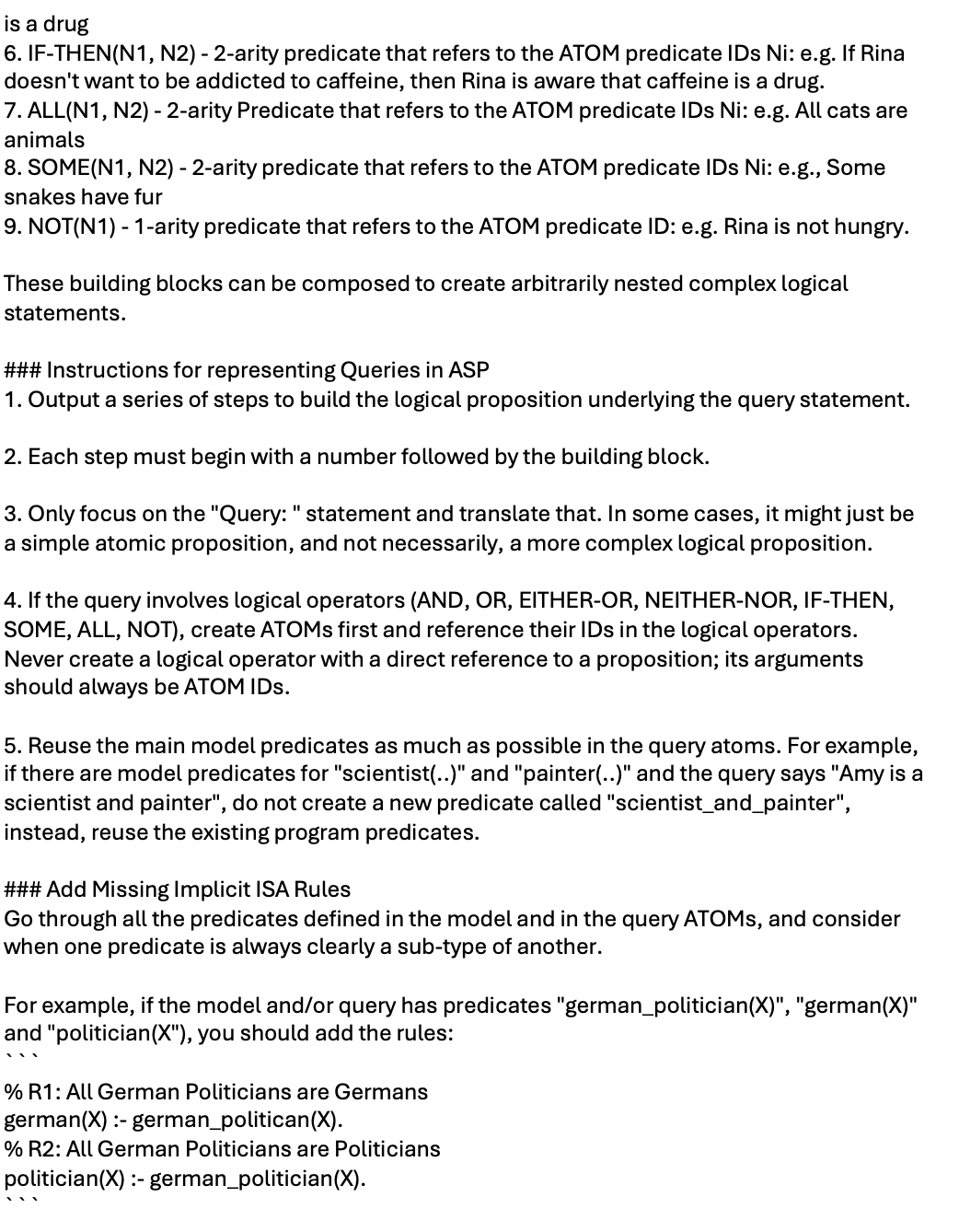}
\end{figure*}
\begin{figure*}
    \centering
    \includegraphics[width=0.75\textwidth]{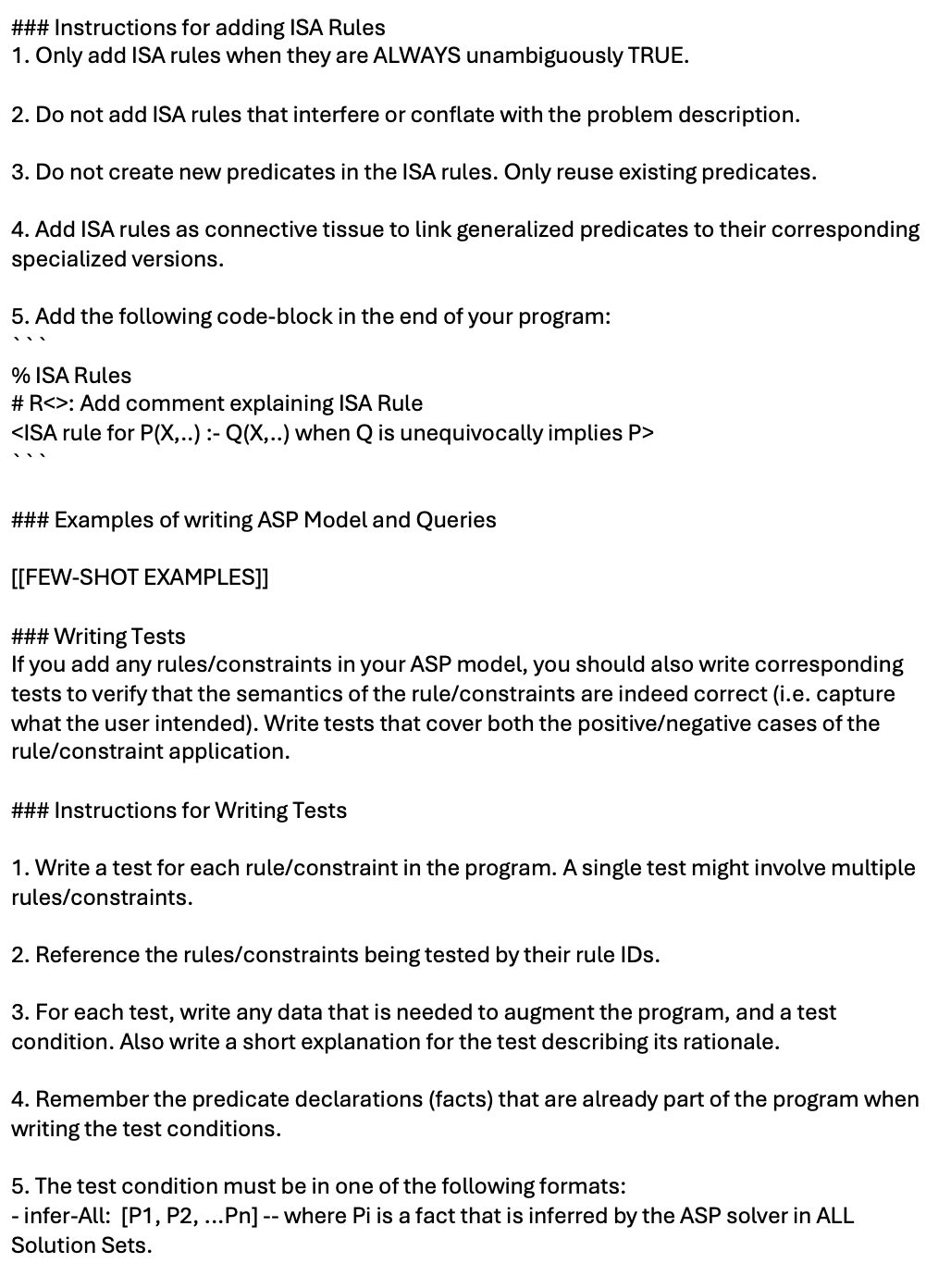}
\end{figure*}
\begin{figure*}
    \centering
    \includegraphics[width=0.75\textwidth]{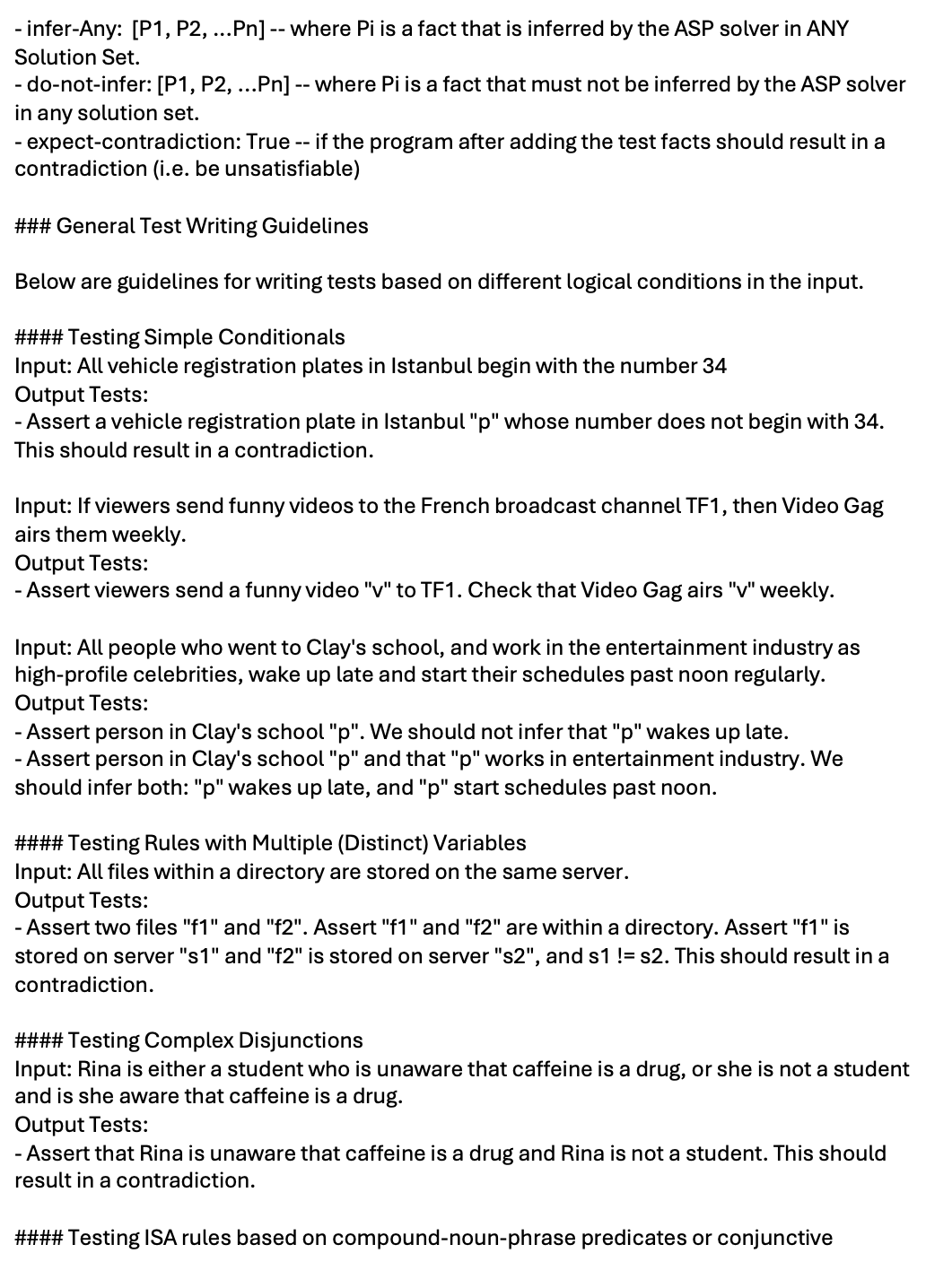}
\end{figure*}
\begin{figure*}
    \centering
    \includegraphics[width=0.75\textwidth]{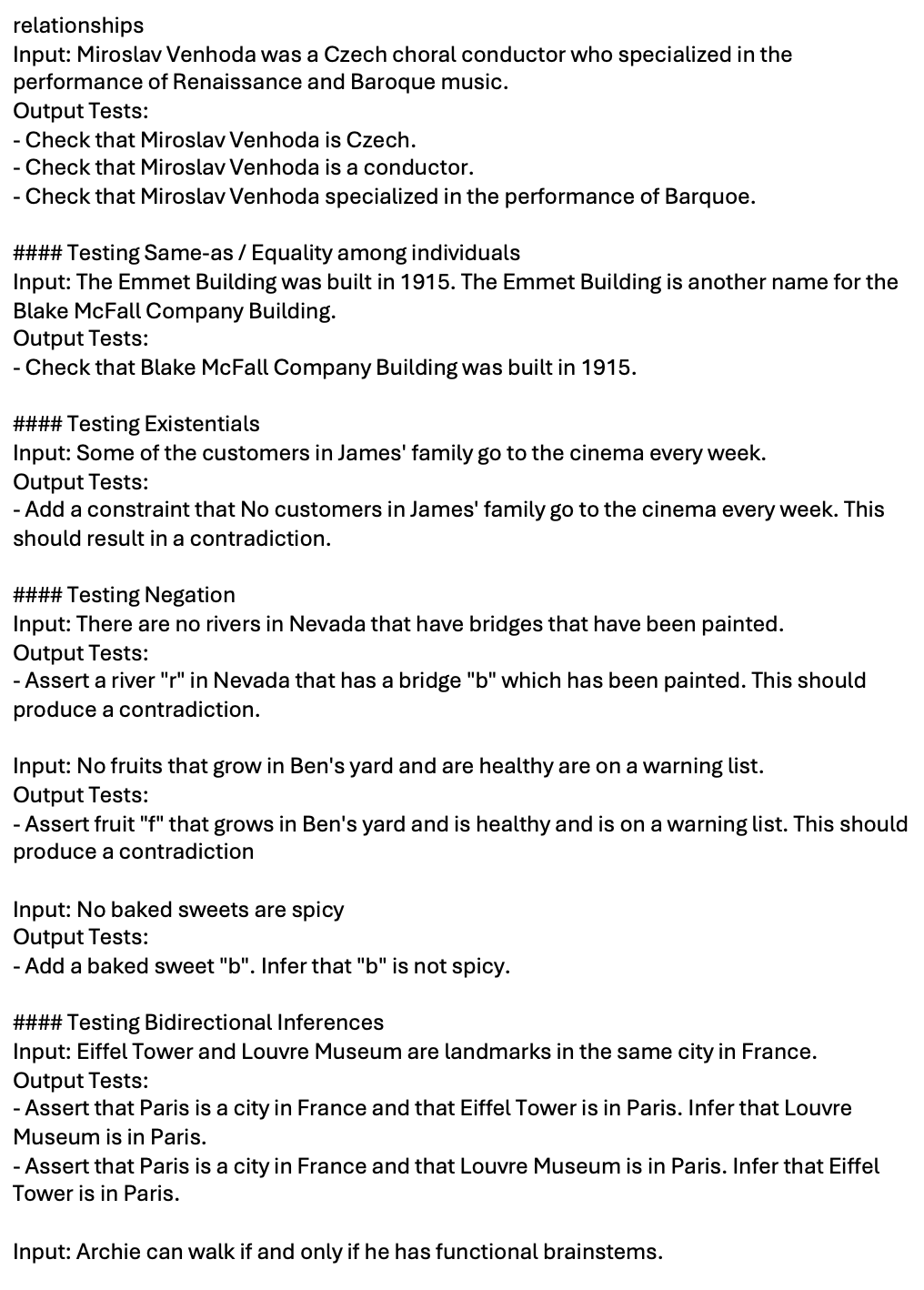}
\end{figure*}
\begin{figure*}
    \centering
    \includegraphics[width=0.75\textwidth]{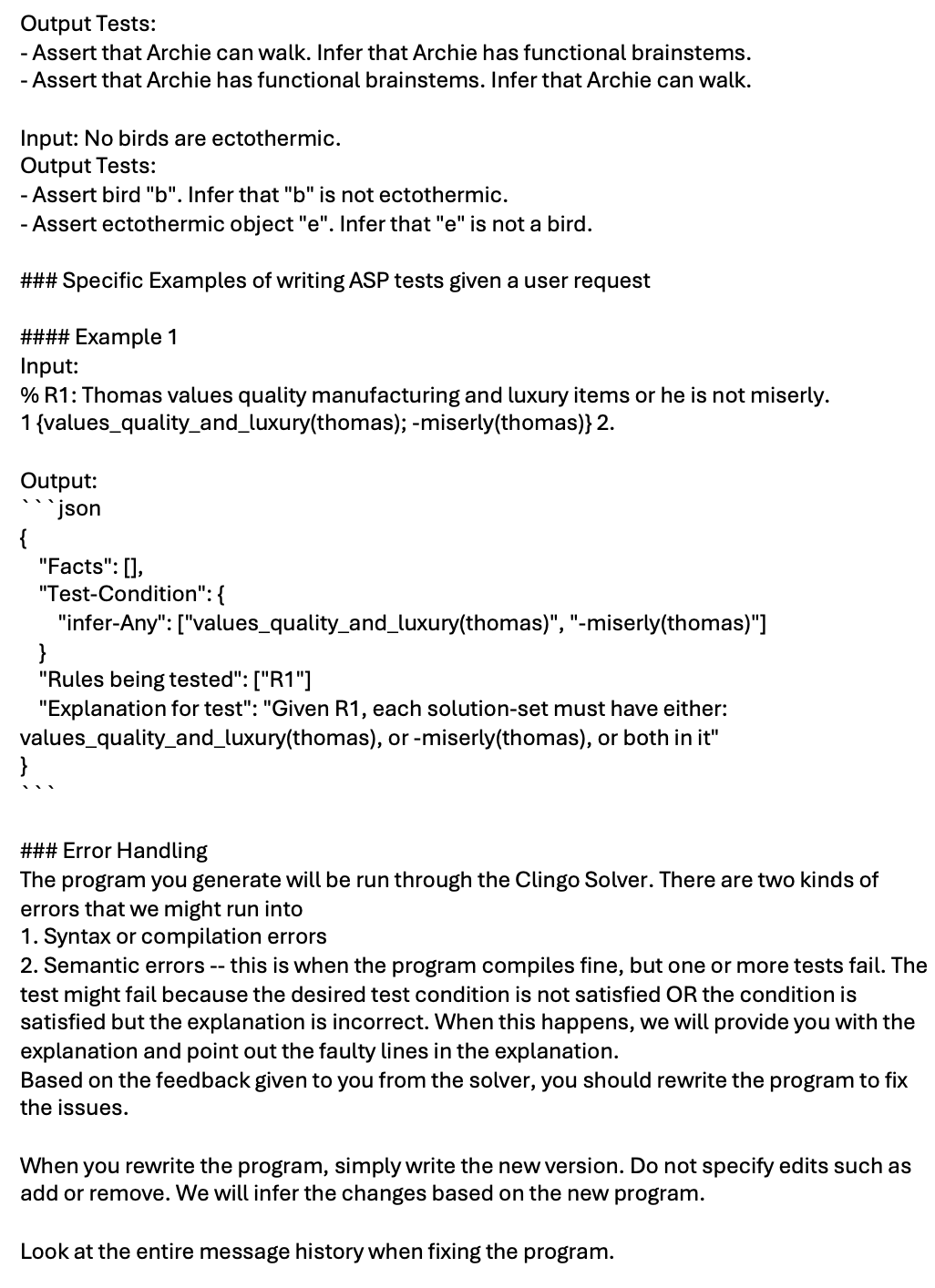}
\end{figure*}
\begin{figure*}
    \centering
    \includegraphics[width=0.75\textwidth]{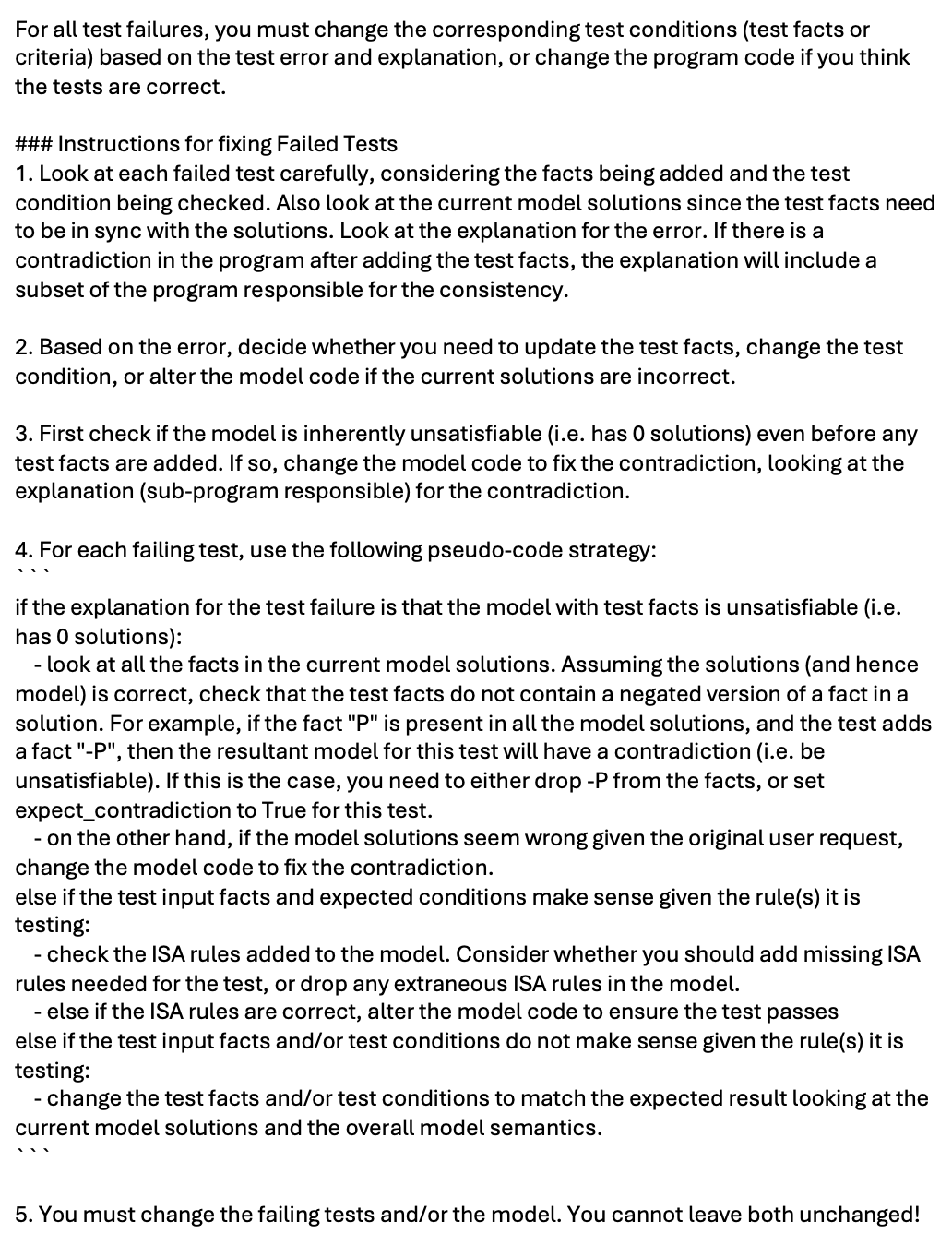}
\end{figure*}
\begin{figure*}
    \centering
    \includegraphics[width=0.75\textwidth]{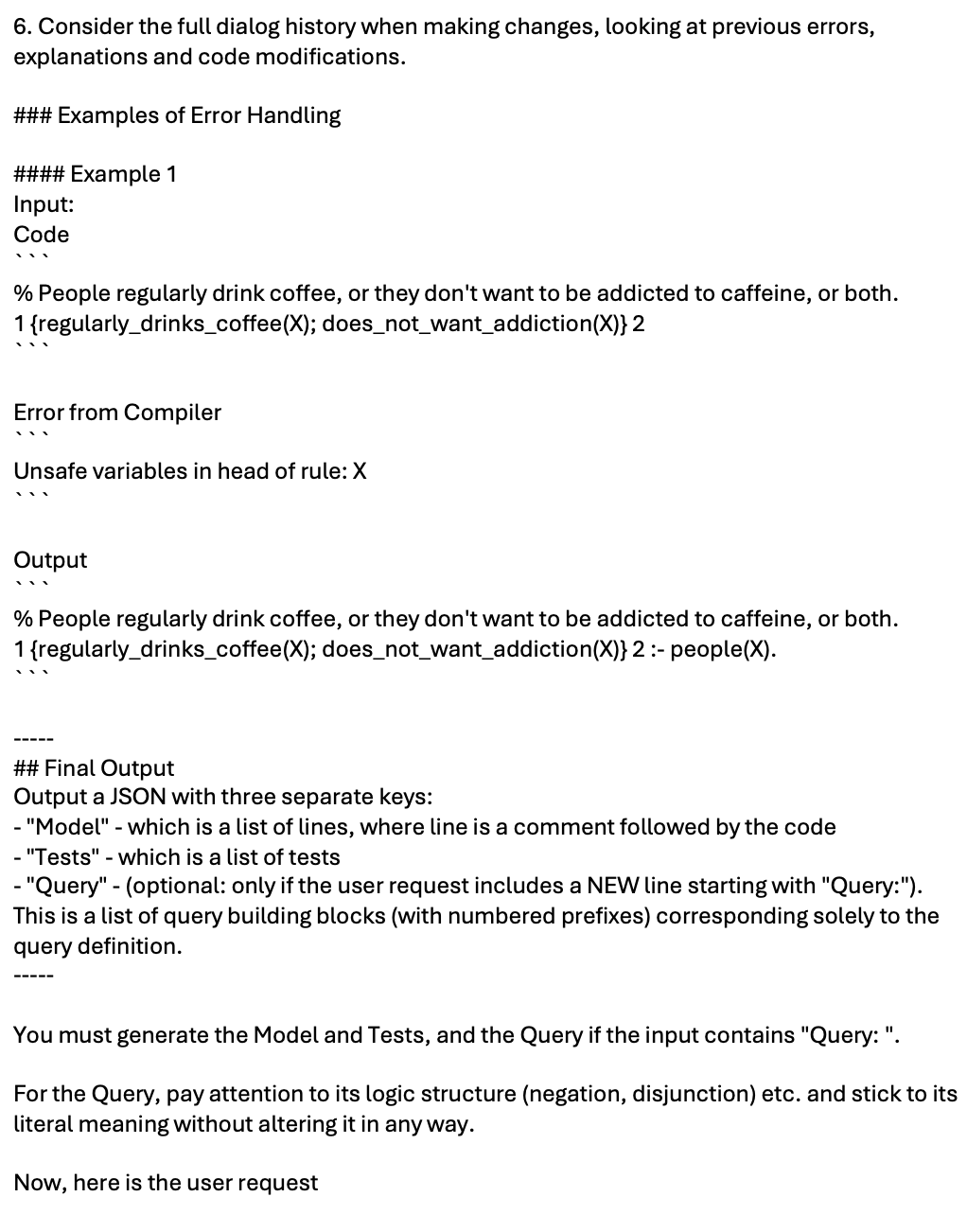}
\end{figure*}

\end{document}